\documentclass{article}

\PassOptionsToPackage{numbers, sort, compress}{natbib}

\usepackage[main,final]{neurips_2026}

\usepackage[utf8]{inputenc} %
\usepackage[T1]{fontenc}    %
\usepackage{hyperref}       %
\usepackage{url}            %
\usepackage{booktabs}       %
\usepackage{amsfonts}       %
\usepackage{amsmath}        %
\usepackage{amssymb}        %
\usepackage{nicefrac}       %
\usepackage{microtype}      %
\usepackage{xcolor}         %
\usepackage{graphicx}       %
\usepackage{wrapfig}        %
\usepackage{algorithm}      %
\usepackage{algpseudocode}  %

\title{PRISM: Primitive Routing via In-context Skill Mixing\\
for Retentive VLA Adaptation}

\author{%
  Jiachen Tao \qquad Gengyu Zhang \qquad Haoxuan Wang\\
  \textbf{Bing Liu \qquad Yan Yan\thanks{Corresponding Author}}\\
  University of Illinois Chicago
}

\begin{document}

\setcounter{footnote}{1}
\maketitle
\setcounter{footnote}{0}

\begin{abstract}
Vision-Language-Action (VLA) models promise a single language-conditioned policy adaptable to many manipulation behaviors, yet deployment additionally requires \emph{skill retention} and \emph{long-horizon composition}: a policy must retain previously acquired skills and reuse them inside composite tasks whose behavior changes across phases. The standard recipe absorbs every new task into one shared update, which provides neither a parameter-level locus for reusable skills nor a mechanism for invoking each skill along the trajectory. We observe that this missing structure is latent rather than absent: per-skill low-rank updates of a pretrained VLA occupy a robustly distinguishable, low-dimensional subspace, and a single composite-task LoRA concentrates roughly half of its Frobenius energy in this subspace, against a near-zero matched random-rotation control. Building on this geometry, we introduce \textbf{PRISM}, which represents reusable skills as parameter-level primitives: each skill family is stored as a dedicated low-rank action-expert adapter in a \emph{primitive bank}, primitives are augmented with skill-specific low-rank residuals and composed at inference by a per-chunk \emph{VLM-conditioned router} that reads the prefix representation to track task progress, and the router is supervised by per-chunk labels from a deterministic, task-conditioned, action-derived oracle. During composite training, the vision encoder, the language LoRA, the skill residuals, the router, and the shared action projections and time MLP are updated; every primitive is held frozen, so the base per-skill action-expert adapters are preserved by construction. On RoboCasa365, PRISM achieves $\mathbf{56.43\%}$ atomic-skill success and reaches $\mathbf{20.00\%}$ on long-horizon composite tasks, while a standard $\pi_{0.5}$-LoRA two-stage fine-tune reaches $17.86\%$ atomic and $17.50\%$ composite success. We additionally validate PRISM on an AgileX PIPER real-robot setup.
\end{abstract}

\section{Introduction}
\label{sec:intro}

Vision-language-action (VLA) models~\citep{brohan2023rt1roboticstransformerrealworld, brohan2023rt2visionlanguageactionmodelstransfer, kim2024openvla, octomodelteam2024octoopensourcegeneralistrobot, black2024pi0visionlanguageactionflowmodel, intelligence2025pi05, nvidia2025gr00tn1openfoundation} are emerging as a general substrate for language-conditioned robot manipulation: a single pretrained policy, steered by natural-language instruction, can in principle be adapted toward a broad range of downstream behaviors. Yet deployment exposes two failure modes of the standard single-policy adaptation paradigm: sequentially fine-tuning the policy on a new behavior degrades performance on previously acquired ones, forcing a choice between retraining from scratch and losing capabilities; and long-horizon tasks that recombine known sub-skills must still be fit end-to-end as if their constituent skills had not already been mastered---an object-washing task, for instance, requires end-to-end training even though it chains a lever-like interaction (to activate a water source) with a pick-and-place-like interaction (to move the object through the stream). We refer to the two requirements that address these failures as \emph{skill retention} and \emph{long-horizon composition}, and ask how a VLA should be parameterized to support both.

The standard adaptation paradigm is monolithic: a downstream task is absorbed into a shared set of model parameters, with no explicit representation of which part of the update corresponds to which reusable behavior. This parameterization may be adequate for mastering one task in isolation, but it lacks two handles required for deployment. The first is \emph{where}: where is a previously learned skill localized in the model update? Without such a parameter-level locus, later updates can interfere with the directions that supported earlier behaviors. Common defenses either allocate additional task-specific parameters, which limits reuse, or constrain updates globally, which can slow acquisition without identifying the skill-specific directions that require protection. The second handle is \emph{when}: when during a long-horizon trajectory should the policy invoke each skill? A composite task cycles through distinct phases of object interaction, motion, and sub-goal. Without a mechanism for switching behavioral regimes, an undifferentiated policy must fit each composite end-to-end as if it were a fresh task. Long-horizon demonstrations are not merely longer action sequences but reflect the selection and recombination of behaviors the policy has already acquired, and a policy with no parameter-level skill identity has no way to interpret such a trajectory in relation to its prior skills.

We find that this missing structure is more latent than absent. When each primitive skill family is adapted on its own through a small low-rank update~\citep{hu2022lora}, the family of per-skill update tensors does not diffuse uniformly throughout parameter space: it occupies a robust, semantically clustered subspace that reproduces across independently trained checkpoints (Section~\ref{sec:prelim:primitive})---low-dimensional relative to the full action-expert parameter space, while remaining nearly full-rank across primitive families. Moreover, when a single LoRA is fit on a composite-task dataset, a substantial fraction of its update tensor lies in the linear span of the per-skill update tensors (Section~\ref{sec:prelim:composite-in-subspace}): composite-task behavior is, to a first approximation, partially expressible as a combination of per-skill primitives. However, when atomic and composite behaviors are absorbed into a single shared adapter trained sequentially, the skill-specific directions learned during atomic adaptation are no longer separately addressable, and atomic-skill performance degrades as composite training proceeds. We refer to this loss of per-skill addressability as \emph{primitive collapse}.

This observation suggests a constructive parameterization. If reusable skills occupy distinguishable low-dimensional update directions, then mixture weights over these updates can be trained to serve as skill-level control variables rather than arbitrary adapter coefficients. Later adaptation need not rewrite a shared update; it can preserve skill-localized slots while learning to route over them. Likewise, long-horizon execution need not fit composite behavior as an independent policy; it can select or mix primitives over time as the task progresses. Under this view, retaining and composing skills reduce to \emph{routing over preserved primitives}: once a skill is represented as a primitive slot, executing a long-horizon behavior activates the appropriate primitive or mixture at each phase.

Building on this observation, we introduce \textbf{PRISM}, a VLA adaptation framework that turns reusable skills into first-class parameter-level objects. PRISM has three coupled components: a \emph{primitive bank} that stores each skill family as its own low-rank adapter over the action expert, augmented with a skill-specific residual during composite training; a \emph{prefix-conditioned router} that emits a per-chunk mixture over these primitive--residual branches; and a \emph{deterministic action-derived labeler} that supplies per-chunk supervision for the router. The bank gives a direct parameter-level answer to the \emph{where} question---adding a new skill amounts to adding a new slot, while every existing primitive remains parametrically untouched.

The router answers the \emph{when} question. At every action chunk it reads the VLM prefix---a contextual feature fusing the current instruction with the current visual state---and emits a distribution over the bank, so the active mixture varies along a long-horizon trajectory. Conditioning on the prefix rather than on the instruction is essential: the instruction is fixed across the trajectory, while sub-goal placement depends on how far the robot has progressed. Training the router from composite demonstrations is nontrivial because behavior cloning alone exerts little pressure for mixture weights to align with skill semantics; PRISM therefore supervises it with per-chunk labels produced by a deterministic, action-derived labeler that requires no learned parameters or per-trajectory human labels once the rule is fixed.

In summary, PRISM makes three contributions. 
\textbf{(i) Structural finding.} Per-skill low-rank updates of VLA action experts span a near-$N$-dimensional, semantically clustered subspace that is reproducible across independently trained banks; a composite-task LoRA places $\sim$$56\%$ of its Frobenius energy in this span, versus $\approx 0\%$ under matched random rotations. 
\textbf{(ii) Framework.} PRISM couples a primitive bank with skill-specific residuals, a per-chunk VLM-conditioned router, and a deterministic action-derived labeler; during composite training, the vision encoder, language LoRA, residuals, router, action projections, and time MLP are updated, while primitives remain frozen.
\textbf{(iii) Evaluation.} On RoboCasa365, PRISM achieves $\mathbf{56.43\%}$ atomic success after composite training and reaches $\mathbf{20.00\%}$ composite success, against $17.86\%$ atomic and $17.50\%$ composite success for a standard $\pi_{0.5}$-LoRA two-stage fine-tune. We further validate PRISM on an AgileX PIPER real robot.

\section{Related Work}
\label{sec:related}

\noindent\textbf{Skill Libraries and Skill Composition.}
\label{sec:related:skills}
PRISM is closest to methods that represent skills as explicit reusable modules. AtomicVLA~\citep{zhang2026atomicvla} routes via skill descriptor tokens; SkillVLA~\citep{zhai2026skillvla} decouples bimanual action experts; PSEC~\citep{liu2025skillexpansioncompositionparameter} composes diffusion-policy LoRA modules in parameter space. PRISM differs in three coupled choices: each skill family is stored as an action-expert LoRA primitive augmented with a trainable residual during composite adaptation; routing is over these branches at action-chunk granularity using the evolving VLM prefix, so the active mixture changes with visual progress rather than with a task or skill identifier; and adapted attention contributions and complete skill-specific FFN outputs are blended in activation space, avoiding cross-products between LoRA factors from different branches. Earlier skill-library and segmentation methods discover, segment, or reuse skills from demonstrations~\citep{strudel2020learningcombineprimitiveskills, wan2024lotuscontinualimitationlearning, mete2024questselfsupervisedskillabstractions, mao2024dexskillsskillsegmentationusing}; PRISM instead assumes a predefined skill taxonomy and tests whether such families are parametrically separable in a pretrained VLA. Concurrent PEFT-library methods DMPEL~\citep{lei2025dmpel} and CLARE~\citep{romer2026clare} grow and route over adapter libraries for lifelong robot learning; PRISM instead targets retention-centric long-horizon composition with chunk-level routing within trajectories rather than acquiring a sequence of discrete tasks.

\noindent\textbf{Parameter-Efficient Fine-Tuning and Mixture of Experts.}
\label{sec:related:peft-moe}
PRISM builds on parameter-efficient adaptation and expert routing. Low-rank adaptation~\citep{hu2022lora}, together with adapter-based and quantized variants~\citep{houlsby2019parameterefficienttransferlearningnlp, dettmers2023qloraefficientfinetuningquantized}, provides a compact interface for adapting large pretrained models without full-model updates. Mixture-of-experts architectures route computation through specialized modules in language and multimodal models~\citep{fedus2022switchtransformersscalingtrillion, jiang2024mixtralexperts, dai2024deepseekmoeultimateexpertspecialization}, and related expert-routing ideas have also been explored in robot policies~\citep{reuss2024efficientdiffusiontransformerpolicies, yu2025forcevlaenhancingvlamodels, huang2025mentormixtureofexpertsnetworktaskoriented}. PRISM treats each primitive--residual branch as a skill expert, but differs from standard MoE in two ways: routing is over skill-localized action-expert adapters at action-chunk granularity, and adapted attention contributions and complete skill-specific FFN outputs are blended in activation space rather than mixing LoRA factors in parameter space.

\noindent\textbf{Continual Learning and Catastrophic Forgetting.}
\label{sec:related:cl}
Continual learning addresses catastrophic forgetting through regularization, rehearsal, and architecture-based expansion. Regularization methods penalize changes to parameters important to past tasks~\citep{kirkpatrick2017overcomingcatastrophicforgettingneural, zenke2017si, li2017lwf}; rehearsal methods replay past examples~\citep{chaudhry2019agem, buzzega2020der}; and architecture-based methods allocate or freeze task-specific parameters~\citep{rusu2016progressive, mallya2018packnet}. In robotics, lifelong learning benchmarks and imitation pipelines study sequential skill acquisition and retention~\citep{liu2023liberobenchmarkingknowledgetransfer, wan2024lotuscontinualimitationlearning}. PRISM is closest in spirit to architecture-based continual learning, but targets a different VLA setting: atomic skills must be retained while being reused inside composite trajectories. Loss- and sample-level defenses such as replay can mitigate forgetting, but they do not provide a skill-localized parameter interface for chunk-level composition. PRISM instead freezes per-skill action-expert primitives during composite training, trains a residual alongside each primitive, and learns to route over these branches, preserving the primitive parameters by construction while evaluating behavioral retention empirically.

\section{Preliminaries and Empirical Skill Geometry}
\label{sec:preliminaries}

\begin{figure}[t]
  \centering
  \includegraphics[width=\linewidth]{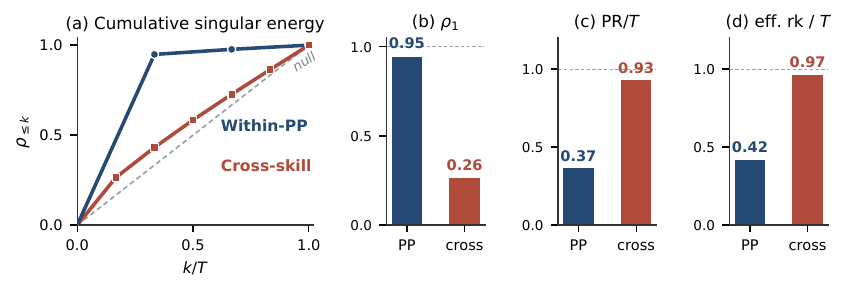}
  \vspace{-25pt}
  \caption{\textbf{Within-skill LoRAs concentrate; across-skill LoRAs spread across $N$.}
  \textbf{(a)} Cumulative singular energy $\rho_{\leq k}$ vs $k/T$: within-PP saturates at the first component, while cross-skill rises along the uniform null $\rho_{\leq k}{=}k/T$ (dashed).
  \textbf{(b)} Top-$1$ energy $\rho_1$: $0.948$ vs $0.265$.
  \textbf{(c)} $\mathrm{PR}/T$ ($1.0{=}$maximal spread, $1/T{=}$collapse): $0.37$ vs $0.93$.
  \textbf{(d)} Effective rank$/T$: $0.42$ vs $0.97$.
  Across all metrics, within-skill LoRAs are more concentrated than across-skill LoRAs.}
  \label{fig:within-skill}
  \vspace{-15pt}
\end{figure}

This section establishes the empirical foundation for PRISM: we define the retention-centric VLA adaptation setting, introduce the LoRA update tensors used for analysis, and show that per-skill updates form a reproducible subspace that captures a substantial fraction of composite-task adaptation.

\subsection{Problem Setup and Notation}
\label{sec:prelim:setup}

\noindent\textbf{Atomic skills and composite tasks.}
We assume access to $N$ \emph{atomic skills} $\mathcal{S} = \{s_1, \ldots, s_N\}$ (skill families at the task level; we use \emph{primitive} for their LoRA-form parametric representation, introduced in §\ref{sec:prelim:perskill}), each paired with demonstrations $D_{s_i}$, and a set of \emph{composite tasks} $\mathcal{T}$ with demonstrations $D_{\text{composite}}$ that cycle through subsets of $\mathcal{S}$ with phase boundaries unlabeled at the chunk level. The retention setting requires that, after composite training on $\mathcal{T}$, per-skill success rate on $\mathcal{S}$ does not degrade; the composition setting requires that performance on $\mathcal{T}$ approaches the per-skill ceiling that ground-truth chunk-level labels would provide.

\noindent\textbf{VLA backbone ($\boldsymbol{\pi_{0.5}}$ substrate).}
We instantiate PRISM on the $\pi_{0.5}$ family~\citep{intelligence2025pi05}, which couples a vision encoder $\phi_\text{V}$, a language model $\phi_\text{L}$, and a small action expert $\pi_\text{A}$ that hosts the per-skill adapters in PRISM. At every chunk boundary $t$, the language transformer processes visual and instruction tokens to condition $\pi_\text{A}$, while the router reads their pooled embedding-level prefix $h_t \in \mathbb{R}^{d_{\text{prefix}}}$ before the language transformer (Equations~\ref{eq:a2-prefix}--\ref{eq:a2-router}). The action expert outputs chunks of $H = 10$ actions, of which $5$ are executed before each re-plan; we use \emph{chunk-level} for any quantity computed once per re-plan.

\noindent\textbf{Adaptation interface.}
The pretrained language-model core and the action-expert backbone $W_{A,0}$ are frozen end to end. Adaptation touches the vision encoder $\phi_\text{V}$ (fully fine-tuned), a shared language LoRA on $\phi_\text{L}$, action projections and time MLP, and per-skill action-expert adapters over $W_{A,0}$: primitives in Phase~1, followed by residuals and the router in Phase~2. Which subset is trainable depends on the training phase (Appendix~\ref{sec:method:training}).

\begin{wrapfigure}{r}{0.45\linewidth}
  \centering
  \vspace{-30pt}
  \includegraphics[width=\linewidth]{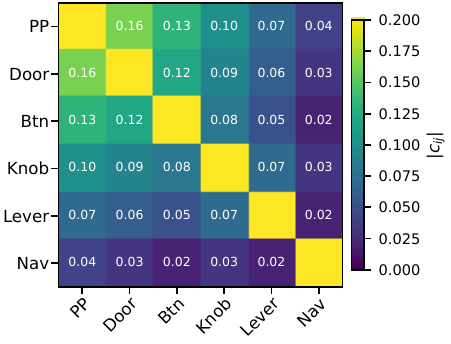}
  \vspace{-16pt}
  \caption{\textbf{Skill geometry, $N{=}6$ bank.}
  Pairwise cosine $c_{ij}$ between layer-wise-flattened per-skill update tensors $\Delta W_i$. Five manipulation skills form a coherent cluster ($0.05$--$0.16$); navigation is a clear outlier ($\le 0.043$).}
  \label{fig:cluster-geometry}
  \vspace{-22pt}
\end{wrapfigure}

\subsection{Per-Skill LoRA Adaptation}
\label{sec:prelim:perskill}

For any layer of the action expert with frozen weight $W_{A,0} \in \mathbb{R}^{d_o \times d_i}$, a rank-$r$ low-rank adaptation~\citep{hu2022lora} augments the layer with a residual $\Delta W = BA$ (so that $W_A = W_{A,0} + \Delta W$), where $B \in \mathbb{R}^{d_o \times r}$ and $A \in \mathbb{R}^{r \times d_i}$. We refer to $\Delta W$ as the layer's \emph{update tensor}. Two protocols are studied. \emph{Per-skill adaptation} jointly trains all pairs $\{(B_i, A_i)\}$ on the union $\bigcup_i D_{s_i}$ under behavior cloning; ground-truth skill labels ensure each batch sample contributes gradients only to its own pair $(B_i, A_i)$, so we recover the per-skill update tensor $\Delta W_i = B_i A_i$. \emph{Composite adaptation} fits a single pair $(B_{\text{comp}}, A_{\text{comp}})$ on $D_{\text{composite}}$ at the same rank, yielding the composite update tensor $\Delta W_{\text{comp}}$.

\subsection{Empirical Skill Geometry}
\label{sec:prelim:primitive}

We measure how the per-skill update tensors $\{\Delta W_i\}_{i=1}^N$ occupy parameter space, and then ask how much of $\Delta W_{\text{comp}}$ lies in their linear span. Measurements come from two independently trained banks ($N{=}6$ and $N{=}11$), allowing us to test whether the resulting geometry reproduces across runs.

\noindent\textbf{Within-skill task collinearity.}
Within a single skill family, per-task LoRAs collapse to a near-$1$D subspace ($\rho_1^{\text{PP}} = 0.948$, $\mathrm{PR}_{\text{PP}} = 1.11$ on a $T{=}3$ Pick~\&~Place pilot), in stark contrast to the cross-skill stack ($\rho_1^{\text{cross}} = 0.265$, $\mathrm{PR}_{\text{cross}} = 5.59 \approx N$); 
see Figure~\ref{fig:within-skill}. This $T{=}3$ pilot provides initial evidence that task variation within a skill family is far more concentrated than variation across skill families; we therefore adopt skill families, rather than individual tasks, as the primitive unit in the present experiments.

\noindent\textbf{Cluster geometry across independently trained checkpoints.}
The pairwise cosines $c_{ij} = \langle \Delta W_i, \Delta W_j \rangle_F / (\|\Delta W_i\|_F \, \|\Delta W_j\|_F)$ on the $6$-skill bank reveal an interpretable manipulation cluster of five end-effector contact skills (Pick~\&~Place, door-opening, knob-turning, button-pressing, lever-turning) with off-diagonal cosines in $[0.05, 0.16]$ and Pick~\&~Place as the hub; navigation (robot base translation) is a clear outlier at $\le 0.043$ against every manipulation skill (Figure~\ref{fig:cluster-geometry}). Across the $15$ skill pairs shared with the $11$-skill bank, the per-pair cosine is preserved with mean $|\Delta c| = 0.016$ and max $|\Delta c| = 0.050$ (on Pick~\&~Place\,$\leftrightarrow$\,door-opening), indicating that the cluster structure reflects task-level motor-primitive families (end-effector contact vs.\ base translation) rather than a single-run optimization artefact.

\begin{wrapfigure}[18]{r}{0.45\linewidth}
  \centering
  \vspace{-7pt}
  \includegraphics[width=\linewidth]{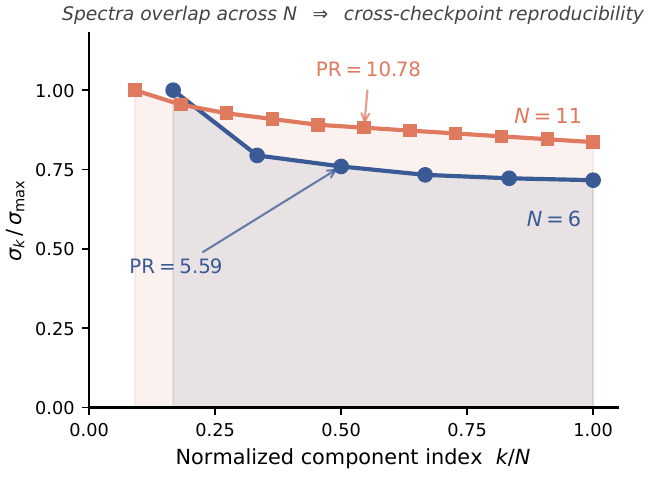}
  \vspace{-16pt}
  \caption{\textbf{Per-skill update tensors span an effectively $N$-dim subspace.} Normalized spectra overlap across $N{=}6$ and $N{=}11$; $\mathrm{PR}{=}5.59$ at $N{=}6$ and $\mathrm{PR}{=}10.78$ at $N{=}11$.}
  \label{fig:primitive-spectrum}
  \vspace{-24pt}
\end{wrapfigure}

\noindent\textbf{Subspace dimensionality.}
Let $G \in \mathbb{R}^{N \times N}$ denote the primitive Gram matrix $G_{ij} = \langle \Delta W_i, \Delta W_j \rangle_F$ with eigenvalues $\{\tau_k\}_{k=1}^N$. The participation ratio
\begin{equation}
\mathrm{PR} \;=\; \frac{\bigl(\sum_{k} \tau_k\bigr)^2}{\sum_{k} \tau_k^2} \;\in\; [1, N]
\label{eq:pr}
\end{equation}
quantifies the number of effectively independent directions the primitives span. We measure $\mathrm{PR} = 5.59$ at $N{=}6$ and $\mathrm{PR} = 10.78$ at $N{=}11$ (Figure~\ref{fig:primitive-spectrum})---in both cases close to the maximum---so adding atomic primitives expands the spanned subspace dimension rather than redundantly populating an already-occupied direction.

\noindent\textbf{Composite--primitive alignment.}
\label{sec:prelim:composite-in-subspace}
Let $\Pi_\text{prim}$ denote the orthogonal projector onto the linear span of $\{\Delta W_i\}_{i=1}^N$. We report the projected energy ratio
\begin{equation}
\eta_\text{comp} \;=\; \frac{\| \Pi_\text{prim} \Delta W_\text{comp} \|_F^2}{\| \Delta W_\text{comp} \|_F^2},
\label{eq:projection-ratio}
\end{equation}

and a control $\eta_\text{rand}$ obtained by replacing each $\Delta W_i$ with a random rotation that preserves Frobenius norm and singular spectrum, averaged over $8$ trials per checkpoint. On both banks, $\eta_\text{comp} \approx 0.56$ vs $\eta_\text{rand} = 0.000 \pm 0.000$ ($0.560$ at $N{=}6$, $0.567$ at $N{=}11$). The alignment is essentially $N$-invariant ($\Delta\eta = 0.007$ across an $83\%$ increase in $N$). Composite-task structure thus lives substantially within the per-skill subspace---a parametric ground for the bank-and-router design in Section~\ref{sec:method}.

\noindent\textbf{Architectural consequence.}
The residual $\sim$$44\%$ may reflect composite-specific structure---most plausibly transition-region dynamics that arise when the policy stitches sub-behaviors together, and that are absent from any single-skill demonstration. This residual lies outside the linear span of the frozen primitives at the parameter level, and is therefore not addressable by routing alone: any time-varying mixture $\sum_i \alpha_i(t)\, B_i A_i$ over the frozen bank still produces an effective action-expert weight within that span. The measured decomposition motivates preserving the primitive updates as a reusable foundation while providing additional trainable directions for composite adaptation. PRISM implements this separation by freezing the primitive bank and adding a skill-specific residual adapter to each primitive. The router selects mixtures of these primitive--residual branches throughout execution, while the vision encoder and language LoRA continue to adapt during composite training (Appendix~\ref{sec:method:training}).

\section{Method}
\label{sec:method}

\begin{figure}[t]
  \centering
  \includegraphics[width=\linewidth]{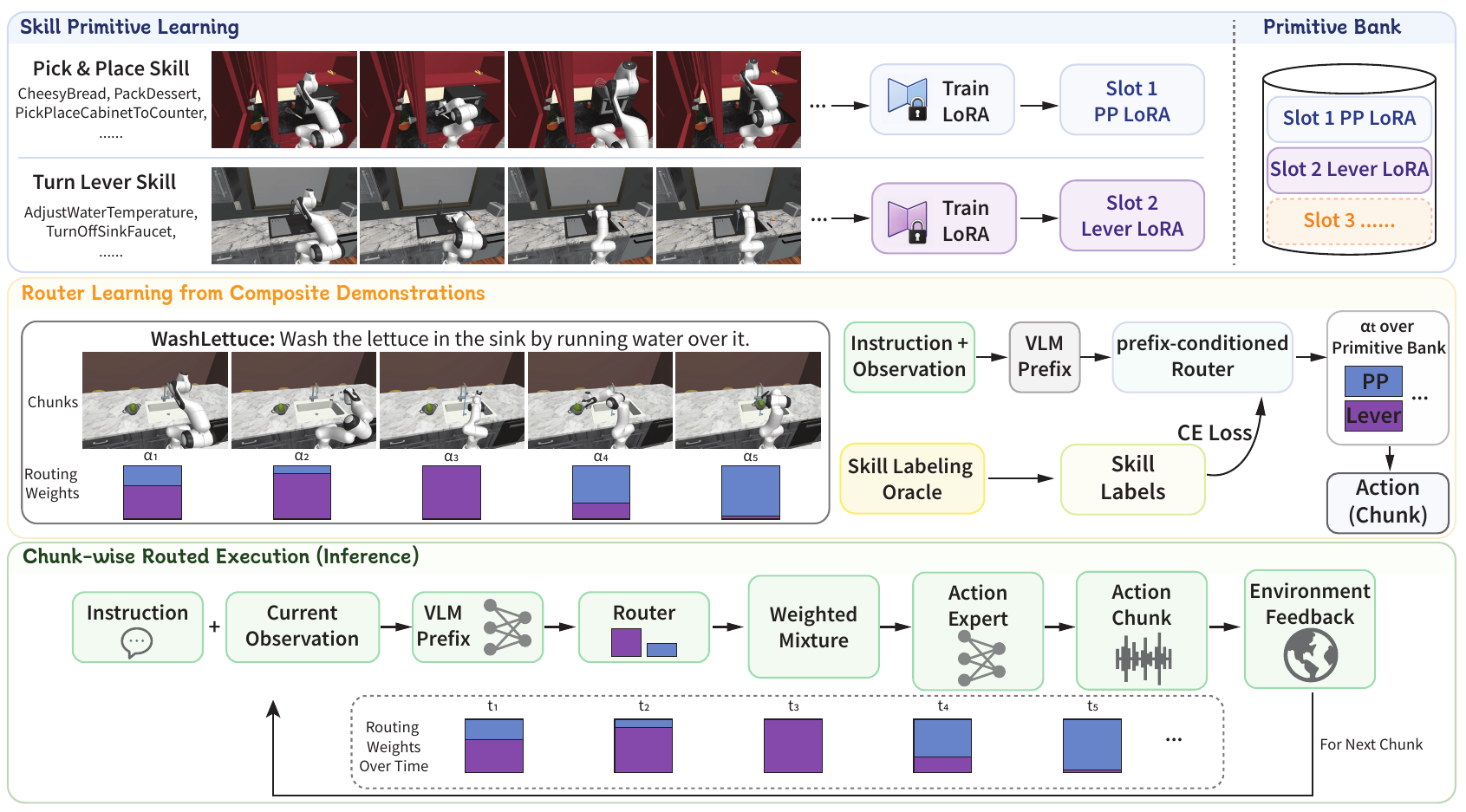}
  \vspace{-20pt}
  \caption{\textbf{Overview of the PRISM framework.}
  \textbf{(1) Skill primitive learning} (§\ref{sec:method:bank}): each atomic skill is fitted with its own rank-$r$ action-expert LoRA $(B_i, A_i)$, yielding a \emph{primitive bank} of disjoint slots, frozen after atomic training.
  \textbf{(2) Router learning from composite demonstrations}: a prefix-conditioned router $r_\theta$ reads the VLM prefix $h_t$ and emits a per-chunk mixture $\alpha(t)$ over the bank, supervised by per-chunk labels from a deterministic action-derived oracle $\mathcal{A}$.
  \textbf{(3) Chunk-wise routed execution} (inference): each LoRA-injected action-expert layer applies the routed mixture at the activation level (Eq.~\ref{eq:a2-attention}), producing an action chunk.}
  \label{fig:pipeline}
  \vspace{-12pt}
\end{figure}

\noindent\textbf{Overview.}
\label{sec:method:overview}
At inference, PRISM takes a language instruction $\ell$ and camera images, and outputs action chunks of $H=10$ actions. It operationalizes the skill geometry of Section~\ref{sec:preliminaries} with three components: a \emph{primitive bank} that stores per-skill LoRA slots augmented with trainable residuals during composite training, a \emph{labeling oracle} that provides skill supervision, and a \emph{VLM-conditioned router} that selects mixtures of the primitive--residual branches. At chunk boundary $t$, the router reads prefix $h_t$ and emits $\alpha(t)=r_\theta(h_t)\in\Delta^{N-1}$. These weights combine adapted attention contributions and complete skill-specific FFN outputs, as defined below. The primitives remain frozen during composite training, while their residuals provide additional adaptation capacity.

\noindent\textbf{Skill Primitives and Residual Adaptation.}
\label{sec:method:bank}
For each atomic skill $s_i$, we learn a rank-$32$ action-expert primitive $P_i=B_iA_i$ from demonstrations $D_{s_i}$ under one-hot routing, so each sample trains its corresponding primitive. The bank $\mathcal{B}=\{(B_i,A_i)\}_{i=1}^{N}$ consists of disjoint parameter blocks: for $i\ne j$, $(B_i,A_i)$ and $(B_j,A_j)$ share no parameters with each other or with the frozen action-expert backbone $W_{A,0}$. The visual encoder and shared language LoRA sit upstream of all primitives. After Phase~1, every primitive is frozen, giving a parameter-level answer to the \emph{where} question: skill $s_i$'s atomic update remains $(B_i,A_i)$.

Composite training introduces a rank-$10$ residual $R_i=D_iC_i$ for every skill, attached to the action expert's attention and feed-forward projections alongside $P_i$. The residual scaling factor is one; $C_i$ is randomly initialized and $D_i$ is initialized to zero. Thus, adding the residual initially preserves each branch's output under the same routing weights. Training subsequently adapts $R_i$ while preserving $P_i$.

\noindent\textbf{Composite Skill Labels.}
\label{sec:method:classifier}
Composite demonstrations provide task-level instructions but no per-chunk atomic-skill labels, so PRISM trains the router with labels from a labeling oracle. Formally, the oracle $\mathcal{A}$ maps a demonstration $(s_1,a_1),\ldots,(s_n,a_n)$ to a skill-index sequence $\{z_t\}_{t=1}^n$, with $z_t\in\{1,\ldots,N\}$, consistent with the atomic skill taxonomy. These labels serve as cross-entropy targets for router training.

We derive frame-level skill labels from task identity and gripper events in demonstration actions. Task identity specifies the ordered skill pair, while gripper opening or closing events anchor the transition between skill phases. Task-dependent temporal offsets include approach and post-release motions in the corresponding phases. These labels provide skill-level supervision for the router; the complete labeling rules are given in Appendix~\ref{app:a2-labels}.

\noindent\textbf{Prefix-Conditioned Routing and Skill Composition.}
\label{sec:method:router}
The router decides, at each chunk boundary, which primitive--residual branches the action expert activates. Its input is a mask-aware mean-pool of the embedding-level prefix tokens:
\begin{equation}
e_t=[\operatorname{SigLIP}(I_t);\,\operatorname{Embed}_{\mathrm{Gemma}}(\ell)],
\label{eq:a2-prefix}
\end{equation}
where $I_t$ denotes the camera observations, SigLIP supplies the visual tokens, and $\operatorname{Embed}_{\mathrm{Gemma}}$ is the instruction-token embedding lookup before the language transformer. With input/padding mask $m_{t,s}$, the router applies LayerNorm and a two-layer MLP followed by softmax:
\begin{equation}
h_t=\frac{\sum_s m_{t,s}e_{t,s}}{\sum_s m_{t,s}},
\qquad
\alpha(t)=\operatorname{softmax}\!\left(\operatorname{MLP}_{\theta}(\operatorname{LN}(h_t))\right).
\label{eq:a2-router}
\end{equation}
The router is initialized to uniform weights and recomputes the mixture for every action chunk.

For each adapted attention projection, the router combines the primitive and residual contributions as
\begin{equation}
y=W_0x+\sum_{i=1}^{N}\alpha_i(t)\bigl(P_ix+R_ix\bigr).
\label{eq:a2-attention}
\end{equation}
For feed-forward layers, each skill branch evaluates its complete FFN, including its nonlinearities, before output mixing:
\begin{equation}
y_{\mathrm{FFN}}=\sum_{i=1}^{N}\alpha_i(t)F_i(x;W_0,P_i,R_i).
\label{eq:a2-ffn}
\end{equation}
Here $F_i$ contains the shared backbone weights and the skill's primitive and residual adapters. Mixing branch outputs preserves each branch's computation and avoids the cross-skill products introduced by blending LoRA factors. The same chunk-level weights govern skill composition throughout the action expert.

\noindent\textbf{Two-Stage Optimization and Execution.}
Phase~1 learns the primitives from atomic demonstrations using task-specified one-hot routing and a flow-matching action objective~\citep{lipman2023flowmatchinggenerativemodeling}. Alongside the primitives, we train the visual encoder, shared language LoRA, action input/output projections, and time MLP; the language and action-expert backbone weights remain frozen. Phase~2 freezes the primitives and trains the residual adapters and router together with these shared adaptation modules on composite demonstrations and atomic replay. The objective is
\begin{equation}
\mathcal{L}_{\mathrm{Phase\,2}}=\mathcal{L}_{\mathrm{FM}}+0.05\,\mathcal{L}_{\mathrm{route}},
\label{eq:a2-loss}
\end{equation}
where $\mathcal{L}_{\mathrm{FM}}$ is the squared error between predicted and target flow velocities, and $\mathcal{L}_{\mathrm{route}}$ is the routing cross-entropy. We mask routing cross-entropy near labeled transitions while retaining the action objective throughout, allowing the policy to learn soft skill mixtures across phase boundaries. Training uses the router's soft weights for action generation, so the action objective and skill supervision jointly optimize routing. Dataset composition and the training schedule are given in Appendix~\ref{sec:method:training}.

At execution time, composite tasks use the learned router to update skill weights for each action chunk. Atomic tasks use task-specified one-hot routing to execute the corresponding primitive--residual branch.

\section{Experiments}
\label{sec:experiments}

We evaluate PRISM in simulation (Sections~\ref{sec:exp:setup}--\ref{sec:exp:ablations}) and on a real robot (Section~\ref{sec:exp:realrobot}). The simulation experiments are designed to answer four questions: (i) does PRISM's primitive-bank-plus-router architecture outperform a standard $\pi_{0.5}$-LoRA fine-tune baseline on long-horizon composite tasks? (ii) does PRISM preserve atomic-skill performance through composite training relative to the same baseline and to a sample-level continual-learning variant (replay)? (iii) which design choices in PRISM (labeling oracle, router conditioning, output composition) actually matter? (iv) do these gains transfer to a physical robot?

\subsection{Simulation Setup}
\label{sec:exp:setup}

\noindent\textbf{Benchmark and skill setup.}
The RoboCasa365~\citep{nasiriany2026robocasa365} atomic dataset contains $3{,}016$ human demonstrations across $28$ tasks: Pick\,\&\,Place ($18$ tasks, $1{,}939$ demonstrations), Turn\,Lever ($3$ tasks, $322$ demonstrations), and Close ($7$ tasks, $755$ demonstrations). These families define the three primitive slots. The composite split contains $415$ demonstrations of LoadDishwasher, SnackSorting, ThawInSink, and WashLettuce. LoadDishwasher and SnackSorting transition from Pick\,\&\,Place to Close; ThawInSink transitions from Pick\,\&\,Place to Turn\,Lever; WashLettuce transitions from Turn\,Lever to Pick\,\&\,Place in the demonstrations. All four tasks require cross-skill composition. Phase~2 combines these demonstrations with $20$ replay episodes per atomic task ($560$ episodes), yielding $975$ episodes and $454{,}758$ frames. Appendix~\ref{app:n3-data} lists the complete task suite and data composition.

\noindent\textbf{Baselines.}
We compare policies built on $\pi_{0.5}$: zero-shot inference, LoRA trained on atomic tasks, LoRA trained sequentially on atomic and composite tasks, and the replay-augmented two-stage LoRA baseline. The PRISM Phase~1 checkpoint isolates atomic primitive learning. Full PRISM adds skill-specific residuals and a per-chunk router during composite training. Table~\ref{tab:n3-main-composite-sr} compares atomic retention and composite success across these configurations.

\subsection{Metrics and Protocol}
\label{sec:exp:metrics}

We report atomic SR as the macro-average over $28$ atomic tasks, with $20$ trials per task, and composite SR as the macro-average over four composite tasks, with $100$ trials per task. PRISM uses task-specified one-hot routing for atomic evaluation and learned per-chunk routing for composite evaluation. The full evaluation protocol is provided in Appendix~\ref{app:eval-protocol}.

\subsection{Main Results: Atomic and Composite Task Success}
\label{sec:exp:main}

\textbf{Higher atomic and composite success.} Table~\ref{tab:n3-main-composite-sr} shows that PRISM achieves $\mathbf{56.43\%}$ atomic success and $\mathbf{20.00\%}$ composite success, exceeding the standard $\pi_{0.5}$-LoRA two-stage fine-tune ($17.86\%$ and $17.50\%$) by $\mathbf{38.57}$ and $\mathbf{2.50}$ percentage points. Relative to LoRA with replay ($49.82\%$ and $10.25\%$), PRISM improves atomic and composite success by $6.61$ and $9.75$ points, respectively, nearly doubling composite success over the replay baseline.

\label{sec:exp:retention}
\textbf{Atomic retention through composite training.} Standard two-stage LoRA fine-tuning reduces atomic success from $48.04\%$ to $17.86\%$, a $30.18$-point drop. PRISM instead raises atomic success from $50.71\%$ in Phase~1 to $56.43\%$ after composite training, a $5.72$-point gain, while keeping the primitive parameters fixed.

\textbf{Composite adaptation of the primitive bank.} PRISM's composite success rises from $0.00\%$ in Phase~1 to $20.00\%$ after Phase~2. Residuals, routing, and shared adaptation modules jointly turn the atomic foundation into a policy that executes the four composite tasks; the ablations isolate the contribution of the routing design.

\begin{table}[t]
  \centering
  \caption{\textbf{Atomic and composite task success rate on RoboCasa365.} Atomic~SR is the macro-average over $28$ atomic tasks ($18$ PP + $3$ Lever + $7$ Close); Composite~SR is the macro-average over $4$ composite tasks. ``Atomic-only'' rows are stage-1 checkpoints; ``atomic + comp.''\ rows are the full two-stage fine-tune. PRISM bank only is the Phase-1 checkpoint without a fitted router. $\Delta$ columns report SR difference (pp) against the $\pi_{0.5}$-LoRA two-stage fine-tune ($\dagger$).}
  \label{tab:n3-main-composite-sr}
  \label{tab:main-composite-sr}
  \small
  \setlength{\tabcolsep}{6pt}
  \begin{tabular}{lcccc}
    \toprule
    Method                                                       & Atomic SR        & $\Delta$ Atomic     & Composite SR     & $\Delta$ Composite  \\
    \midrule
    $\pi_{0.5}$ (zero-shot)                                      & $0.00$ & $-17.86$ & $0.00$ & $-17.50$ \\
    $\pi_{0.5}$-LoRA, atomic-only                                & $48.04$ & $+30.18$ & $0.00$ & $-17.50$ \\
    $\pi_{0.5}$-LoRA, atomic + comp.\ $^{\dagger}$               & $17.86$ & $0.00$ & $17.50$ & $0.00$ \\
    $\pi_{0.5}$-LoRA + replay                                    & $49.82$ & $+31.96$ & $10.25$ & $-7.25$ \\
    \midrule
    PRISM, bank only (Phase~1)                                   & $50.71$ & $+32.85$ & $0.00$ & $-17.50$ \\
    \textbf{PRISM (ours)}                                        & $\mathbf{56.43}$ & $+38.57$ & $\mathbf{20.00}$ & $+2.50$ \\
    \bottomrule
  \end{tabular}
  \vspace{-8pt}
\end{table}

\subsection{Ablations}
\label{sec:exp:ablations}

We ablate three core PRISM design choices in Table~\ref{tab:ablation-panel}: (a) which oracle provides per-chunk skill labels for router training, (b) what input the router conditions on, and (c) how the router output is composed with the primitive bank. Each panel reports composite macro-average SR.

\begin{table}[t]
  \centering
  \caption{\textbf{Three ablations on PRISM design choices, reported as composite macro-average SR.} (a) The composite skill labeling oracle of Section~\ref{sec:method:classifier}. (b) The router conditioning input (Section~\ref{sec:method:router}). (c) The router output and primitive composition mechanism.}
  \label{tab:ablation-panel}
  \footnotesize
  \begin{minipage}[t]{0.32\linewidth}
    \centering
    \textbf{(a) Labeling oracle}\\[3pt]
    \setlength{\tabcolsep}{4pt}
    \begin{tabular}{lc}
      \toprule
      Oracle                  & Comp.\ Avg       \\
      \midrule
      Qwen2.5-VL              & $11.25$          \\
      MLP $(s_t, a_t)$        & $13.75$          \\
      \textbf{Kinematic (ours)} & $\mathbf{20.00}$ \\
      \bottomrule
    \end{tabular}
  \end{minipage}%
  \hfill
  \begin{minipage}[t]{0.32\linewidth}
    \centering
    \textbf{(b) Router input}\\[3pt]
    \setlength{\tabcolsep}{4pt}
    \begin{tabular}{lc}
      \toprule
      Input                   & Comp.\ Avg       \\
      \midrule
      Vision only             & $10.25$          \\
      State only              & $7.50$           \\
      \textbf{VLM prefix (ours)} & $\mathbf{20.00}$ \\
      \bottomrule
    \end{tabular}
  \end{minipage}%
  \hfill
  \begin{minipage}[t]{0.32\linewidth}
    \centering
    \textbf{(c) Router output \& composition}\\[3pt]
    \setlength{\tabcolsep}{4pt}
    \begin{tabular}{lc}
      \toprule
      Composition             & Comp.\ Avg       \\
      \midrule
      Hard one-hot            & $10.50$          \\
      Soft, param-space       & $15.50$          \\
      Soft, action-space      & $13.50$          \\
      \textbf{Soft, activation (ours)} & $\mathbf{20.00}$ \\
      \bottomrule
    \end{tabular}
  \end{minipage}
  \vspace{-10pt}
\end{table}

\noindent\textbf{Labeling oracle (panel a).}
\label{sec:exp:abl:oracle}
The oracle $\mathcal{A}$ is a swappable interface (§\ref{sec:method:classifier}); we compare our default kinematic rule against a Qwen2.5-VL~\citep{bai2025qwen25vltechnicalreport} oracle on video frames and a learned MLP on $(s_t, a_t)$. The kinematic rule outperforms both ($\mathbf{20.00}$ vs MLP $13.75$ vs VLM $11.25$): the VLM\textquotesingle s transition placement varies under prompt and viewpoint changes, while the deterministic rule anchors skill transitions to task identity and gripper events.

\noindent\textbf{Router conditioning (panel b).}
\label{sec:exp:abl:router-input}
A chunk-level routing decision needs both task identity and progress within the task. Pooled vision features carry progress but no instruction identity ($10.25$); proprioceptive state is too low-dimensional to track phase transitions ($7.50$); only the VLM prefix $h_t$ fuses both signals into a single contextual feature ($\mathbf{20.00}$).

\noindent\textbf{Router output and composition (panel c).}
\label{sec:exp:abl:router-output}
\emph{Hard one-hot} ($10.50$) imposes discrete branch selection across the cross-skill transitions.

\emph{Soft + parameter-space} ($15.50$) mixes LoRA factors before applying the update. For a set of factors $(B_i,A_i)$, the schematic product $(\sum_i\alpha_i B_i)(\sum_j\alpha_j A_j)$ introduces cross-skill terms $\alpha_i\alpha_j B_iA_j$ for $i\ne j$.

\emph{Soft + action-space} ($13.50$) executes $N$ policy forward passes and averages the resulting action chunks at the output.

\emph{Soft + activation-space} (PRISM, $\mathbf{20.00}$) blends primitive--residual contributions in attention projections and complete skill-specific FFN outputs, avoiding cross-skill LoRA-factor products within one action-expert forward pass.

\subsection{Real-Robot Experiments}
\label{sec:exp:realrobot}

We evaluate PRISM on an AgileX PIPER 6-DoF arm with two atomic skills (Cup\,$\rightarrow$\,Table, Doll\,$\rightarrow$\,Plate) and one composite chaining them; gripper events in demonstration actions provide skill-phase supervision (Section~\ref{sec:method:classifier}). We collect 20 teleoperated demonstrations per task and roll out 20 trials per task with matched initial configurations; full hardware, task, and protocol details are in Appendix~\ref{app:eval-protocol}. Figure~\ref{fig:real-robot-comp} shows a successful PRISM rollout on the composite task.

\begin{figure}[t]
  \centering
  \includegraphics[width=\linewidth]{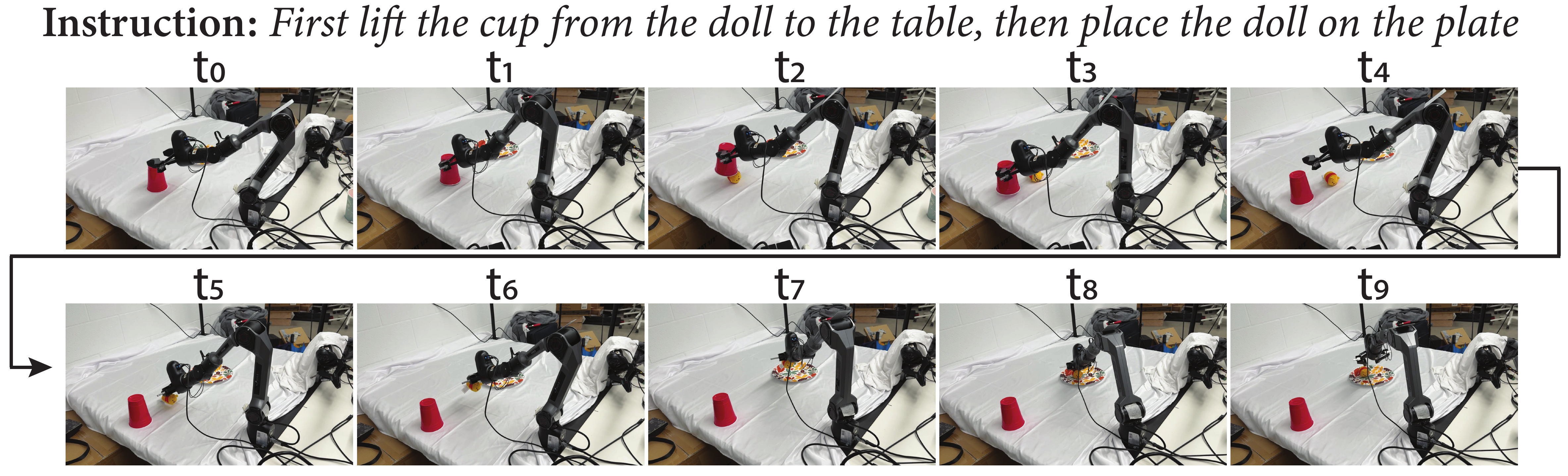}
  \vspace{-10pt}
  \caption{\textbf{Real-robot composite rollout under PRISM.} A successful inference on the Cup\,$\rightarrow$\,Table\,$\rightarrow$\,Doll\,$\rightarrow$\,Plate composite task: the robot first lifts the cup off the doll ($t_0$--$t_4$), then picks up the doll and places it on the plate ($t_5$--$t_9$). The router supplies a per-chunk mixture $\alpha(t)$ over the bank that switches from the Cup\,$\rightarrow$\,Table primitive to the Doll\,$\rightarrow$\,Plate primitive mid-trajectory.}
  \label{fig:real-robot-comp}
  \vspace{-5pt}
\end{figure}

\begin{table}[t]
  \centering
  \caption{\textbf{Real-robot success rate on AgileX PIPER.} Two atomic skills (Cup\,$\rightarrow$\,Table, Doll\,$\rightarrow$\,Plate) and a Composite chaining them. $\Delta$ columns report absolute SR difference (pp) against the $\pi_{0.5}$-LoRA two-stage fine-tune ($\dagger$). SR$\uparrow$, $\Delta\uparrow$.}
  \label{tab:real-robot-sr}
  \small
  \setlength{\tabcolsep}{5pt}
  \begin{tabular}{lcccccc}
    \toprule
    & \multicolumn{2}{c}{Cup$\rightarrow$Table} & \multicolumn{2}{c}{Doll$\rightarrow$Plate} & \multicolumn{2}{c}{Composite} \\
    \cmidrule(lr){2-3} \cmidrule(lr){4-5} \cmidrule(lr){6-7}
    Method                                              & SR              & $\Delta$           & SR              & $\Delta$           & SR              & $\Delta$           \\
    \midrule
    $\pi_{0.5}$ (zero-shot)                             & $0.0$           & $-25.0$            & $0.0$           & $-35.0$            & $0.0$           & $-45.0$            \\
    $\pi_{0.5}$-LoRA, atomic-only                       & $55.0$          & $+30.0$            & $75.0$          & $+40.0$            & $0.0$           & $-45.0$            \\
    $\pi_{0.5}$-LoRA, atomic + comp.\ $^{\dagger}$      & $25.0$          & $\phantom{+}0$     & $35.0$          & $\phantom{+}0$     & $45.0$          & $\phantom{+}0$     \\
    $\pi_{0.5}$-LoRA + replay                           & $50.0$          & $+25.0$            & $65.0$          & $+30.0$            & $50.0$          & $+5.0$             \\
    \midrule
    PRISM, bank only (Phase~1)                          & $65.0$          & $+40.0$            & $80.0$          & $+45.0$            & $0.0$           & $-45.0$            \\
    \textbf{PRISM (ours)}                               & $\mathbf{65.0}$ & $\mathbf{+40.0}$   & $\mathbf{85.0}$ & $\mathbf{+50.0}$   & $\mathbf{55.0}$ & $\mathbf{+10.0}$   \\
    \bottomrule
  \end{tabular}
  \vspace{-7pt}
\end{table}

\noindent\textbf{Results.}
Table~\ref{tab:real-robot-sr} mirrors the simulation pattern. The $\pi_{0.5}$-LoRA two-stage fine-tune loses $30$--$40$ atomic points through composite training; replay partially recovers atomic but still trails on composite. PRISM preserves atomic SR through Phase~2 (Cup\,$\rightarrow$\,Table $65\%{\to}65\%$, Doll\,$\rightarrow$\,Plate $80\%{\to}85\%$) and reaches $\mathbf{55.0\%}$ composite---highest in every column.

\section{Conclusion}
\label{sec:conclusion}

We presented PRISM, a VLA adaptation framework that represents reusable skills as parameter-level primitives. PRISM exploits an empirical structural finding---per-skill LoRA updates span a near-$N$-dimensional, semantically clustered subspace that captures $\sim$$56\%$ of a composite-task LoRA's energy---and operationalizes it through a primitive bank with skill-specific residual adaptation, a per-chunk VLM-conditioned router, and a deterministic action-derived labeler. On RoboCasa365, PRISM achieves $\mathbf{56.43\%}$ atomic success after composite training and reaches $\mathbf{20\%}$ on long-horizon composite tasks, outperforming both two-stage fine-tuning and replay. Sequential addition of new skill families and scaling beyond three primitives are natural next steps for the bank-plus-router design.

\begin{ack}
This research was partially supported by NSF grants IIS-2525840, CNS-2432534, and ECCS-2514574. Bing Liu's work was supported in part by NSF grants IIS-2229876 and CNS-2225427. The views and conclusions expressed in this article are solely those of the authors and do not necessarily reflect those of the funding agencies or supporting organizations.
\end{ack}

\bibliographystyle{plainnat}
\bibliography{refs}

\appendix

\section{Training Pipeline}
\label{sec:method:training}

\paragraph{Backbone configuration.}
PRISM is built on $\pi_{0.5}$~\citep{intelligence2025pi05}: vision encoder $\phi_\text{V}$ is the SigLIP encoder~\citep{zhai2023siglip} from PaliGemma~\citep{beyer2024paligemmaversatile3bvlm}; language model $\phi_\text{L}$ is Gemma~\citep{gemmateam2024gemma} with a shared rank-$16$ LoRA $L$ on attention and feed-forward sublayers; action expert $\pi_\text{A}$ is Gemma, into which three per-skill primitive LoRAs (rank-$32$) are injected over the action-expert weight $W_{A,0}$. In Phase~2, each primitive is augmented with a trainable rank-$10$ residual.

PRISM is trained by a two-phase pipeline.

\paragraph{Phase~1 (atomic primitive learning, $T_1 = 30$K steps).}
Starting from pretrained $\pi_{0.5}$ weights, each atomic demonstration is hard-assigned to its skill slot $i$ (one-hot $\alpha$), routing the whole episode through $(B_i,A_i)$ alone. We train the bank $\{(B_i,A_i)\}_{i=1}^{3}$ through the selected primitive, full-fine-tune $\phi_\text{V}$, and adapt the shared rank-$16$ language LoRA $L$, action input/output projections, and time MLP. Both Gemma backbones, language token embeddings, and language normalization layers are frozen. The atomic dataset is described in Appendix~\ref{app:n3-data}. Training uses batch size $32$, $1{,}000$ warmup steps, and a cosine learning-rate schedule from $2.5\times10^{-5}$ to $2.5\times10^{-6}$. Phase~2 starts from the final Phase-1 checkpoint.

\paragraph{Phase~2 (joint composite training, $T_2 = 20$K steps).}
We warm-start from Phase~1 and train on $D_\text{replay}$, which combines $415$ composite demonstrations with $K=20$ replay episodes per atomic task ($560$ episodes across $28$ tasks; Appendix~\ref{app:n3-data}). Frames are sampled uniformly from the merged dataset, without class balancing. Trainable parameters are $\phi_\text{V}$, $L$, $r_\theta$, the three rank-$10$ skill residuals, action input/output projections, and time MLP; the primitive bank and both Gemma backbones remain frozen.

With $z_t$ from the labeling oracle (Section~\ref{sec:method:classifier}), the loss is $\mathcal{L}_{\mathrm{FM}}+0.05\mathcal{L}_{\mathrm{route}}$ (Equation~\ref{eq:a2-loss}). Routing cross-entropy excludes frames within $100$ frames on either side of a labeled transition, while the flow-matching objective uses all sampled frames. Action prediction uses soft router weights, so the action objective also updates the router. The routing loss propagates into $\phi_\text{V}$ through the embedding-level prefix; $L$ receives gradients through the action objective.

Training uses batch size $32$, $500$ warmup steps, and a cosine learning-rate schedule from $2.5\times10^{-5}$ to $2.5\times10^{-6}$. Evaluation uses the final Phase-2 checkpoint.

\paragraph{Why freeze the bank in Phase~2.}
Jointly updating the primitive bank allows composite training to modify the atomic updates, breaking the parameter-isolation contract. Freezing the bank preserves the atomic updates while skill-specific residuals and shared modules adapt to composite behavior. The router combines the primitive--residual branches, while $\mathcal{L}_{\text{CE}}$ against $z_t$ anchors $r_\theta$ to the three skill families. Replay atomic frames provide single-skill examples that anchor the router on the atomic distribution, while composite frames train the residuals, router, and shared modules on phase-switching trajectories.

\paragraph{Lifelong extension.}
The bank supports adding a new skill through an additional primitive--residual branch and a corresponding router output, while preserving the existing primitive parameters.

Existing primitives are never updated after their own Phase-1 training, so the bank's frozen-after-Phase-1 property is preserved across additions.

\section{Limitations}
\label{sec:limitations}

\paragraph{Skill taxonomy and supervision.}
PRISM uses predefined skill families, and its action-derived labeling rules use task identity, gripper events, and temporal offsets to assign skill phases. These rules assume that the relevant transitions can be localized from demonstration actions; tasks without such event cues require a different labeling mechanism. The router-supervision interface accommodates alternative labelers, including the learned classifier and VLM evaluated in Section~\ref{sec:exp:abl:oracle}.

\paragraph{Evaluation scope and retention.}
The policy evaluation studies atomic-skill retention through composite adaptation with a fixed bank: three primitives and four composite tasks in simulation, and two primitives and one composite task on the real robot. The real-robot evaluation uses closely matched initial configurations. The independent $N=6$ and $N=11$ banks in Section~\ref{sec:preliminaries} establish the measured primitive geometry; sequential skill acquisition and policy evaluation with larger banks remain extensions of the present setting. Freezing the primitive bank preserves its parameters, while skill residuals and shared modules continue to adapt. Behavioral retention is therefore measured by atomic-task success after composite training, using task-specified routing; composite evaluation uses the learned router.

\section{Evaluation Protocol}
\label{app:eval-protocol}

This appendix documents the evaluation protocol in detail, complementing the headline metrics described in Section~\ref{sec:exp:metrics}.

\paragraph{Per-task rollouts.}
The RoboCasa365 atomic suite uses $20$ trials per task and the composite suite uses $100$ trials per task. PRISM uses seed base $17$ for atomic evaluation and $31$ for composite evaluation. The physical-robot study uses $20$ trials per task.

\paragraph{Success criterion.}
For RoboCasa365 atomic and composite tasks, success is determined by the simulator's built-in task-completion checker. For the real-robot setup, success is judged by a human evaluator (Cup\,$\rightarrow$\,Table: cup placed on the table surface without dropping; Doll\,$\rightarrow$\,Plate: doll placed on the plate without dropping; Composite: both succeed in sequence).

\paragraph{Aggregation.}
Atomic SR averages success equally over $28$ tasks, and composite SR averages success equally over four tasks. Equal trial counts within each suite make these macro-averages equal to the pooled success fractions.

\paragraph{Execution modes.}
Atomic evaluation uses task-specified routing, activating the corresponding primitive for Phase~1 and the primitive--residual branch for the complete policy. Composite evaluation of the complete policy recomputes the learned router weights at each replanning step.

\subsection{RoboCasa365 Dataset Details}
\label{app:robocasa-dataset}
\label{app:n3-data}

\paragraph{Dataset summary.}
All training demonstrations are human demonstrations from RoboCasa365, recorded at $20$\,Hz. The atomic split contains $3{,}016$ demonstrations and $687{,}997$ frames across $28$ tasks in three skill families: Pick\,\&\,Place, Turn\,Lever, and Close. Table~\ref{tab:robocasa-atomic-n3} gives the complete inventory. LoadDishwasher, SnackSorting, ThawInSink, and WashLettuce supply $415$ composite demonstrations and $327{,}608$ frames.

\begin{table}[h]
  \centering
  \caption{\textbf{RoboCasa365 atomic dataset.} Per-task episode and frame counts for the $28$ atomic tasks used in PRISM's training and evaluation.}
  \label{tab:robocasa-atomic-n3}
  \footnotesize
  \setlength{\tabcolsep}{8pt}
  \begin{tabular}{lrr}
    \toprule
    Task                                  & Episodes & Frames \\
    \midrule
    \multicolumn{3}{l}{\textbf{Pick \& Place} ($18$ tasks)} \\
    \quad PickPlaceCabinetToCounter       & $106$ & $20{,}201$ \\
    \quad PickPlaceCounterToBlender       & $106$ & $38{,}892$ \\
    \quad PickPlaceCounterToCabinet       & $108$ & $24{,}225$ \\
    \quad PickPlaceCounterToDrawer        & $107$ & $28{,}225$ \\
    \quad PickPlaceCounterToMicrowave     & $110$ & $42{,}012$ \\
    \quad PickPlaceCounterToOven          & $111$ & $32{,}014$ \\
    \quad PickPlaceCounterToSink          & $108$ & $22{,}410$ \\
    \quad PickPlaceCounterToStandMixer    & $107$ & $25{,}467$ \\
    \quad PickPlaceCounterToStove         & $108$ & $24{,}039$ \\
    \quad PickPlaceCounterToToasterOven   & $108$ & $24{,}313$ \\
    \quad PickPlaceDrawerToCounter        & $103$ & $31{,}819$ \\
    \quad PickPlaceFridgeDrawerToShelf    & $106$ & $26{,}396$ \\
    \quad PickPlaceFridgeShelfToDrawer    & $111$ & $27{,}047$ \\
    \quad PickPlaceMicrowaveToCounter     & $112$ & $38{,}729$ \\
    \quad PickPlaceSinkToCounter          & $108$ & $26{,}397$ \\
    \quad PickPlaceStoveToCounter         & $109$ & $23{,}003$ \\
    \quad PickPlaceToasterOvenToCounter   & $106$ & $19{,}323$ \\
    \quad PickPlaceToasterToCounter       & $105$ & $26{,}907$ \\
    \cmidrule(lr){1-3}
    \quad \emph{subtotal}                 & \emph{$1{,}939$} & \emph{$501{,}419$} \\
    \midrule
    \multicolumn{3}{l}{\textbf{Turn Lever} ($3$ tasks)} \\
    \quad TurnOnSinkFaucet                & $107$ & $23{,}795$ \\
    \quad TurnOffSinkFaucet               & $106$ & $12{,}309$ \\
    \quad TurnSinkSpout                   & $109$ & $11{,}130$ \\
    \cmidrule(lr){1-3}
    \quad \emph{subtotal}                 & \emph{$322$} & \emph{$47{,}234$} \\
    \midrule
    \multicolumn{3}{l}{\textbf{Close} ($7$ tasks)} \\
    \quad CloseCabinet & $105$ & $27{,}754$ \\
    \quad CloseDishwasher & $111$ & $15{,}781$ \\
    \quad CloseDrawer & $110$ & $15{,}670$ \\
    \quad CloseFridge & $106$ & $26{,}888$ \\
    \quad CloseFridgeDrawer & $111$ & $14{,}946$ \\
    \quad CloseMicrowave & $105$ & $20{,}075$ \\
    \quad CloseOven & $107$ & $18{,}230$ \\
    \cmidrule(lr){1-3}
    \quad \emph{subtotal} & \emph{$755$} & \emph{$139{,}344$} \\
    \midrule
    \textbf{Total} & $\mathbf{3{,}016}$ & $\mathbf{687{,}997}$ \\
    \bottomrule
  \end{tabular}
\end{table}

\paragraph{Composite demonstrations.}
LoadDishwasher requires placing a cup and bowl in the dishwasher and closing its door; both placement interactions belong to the Pick\,\&\,Place phase. SnackSorting requires placing a bar in a bowl and closing a drawer. ThawInSink requires placing the instructed object in the sink and then turning on the faucet. WashLettuce asks the robot to wash lettuce with running water; its demonstration-labeling rule assigns Turn\,Lever followed by Pick\,\&\,Place. Task identity specifies the skill order, and gripper events locate the transition.

\begin{center}
\small
\begin{tabular}{lrrr}
\toprule
Composite task & Episodes & Frames & Mean frames / episode \\
\midrule
LoadDishwasher & $100$ & $84{,}383$ & $843.83$ \\
SnackSorting & $100$ & $82{,}518$ & $825.18$ \\
ThawInSink & $111$ & $81{,}459$ & $733.86$ \\
WashLettuce & $104$ & $79{,}248$ & $762.00$ \\
\midrule
Total & $415$ & $327{,}608$ & $789.42$ \\
\bottomrule
\end{tabular}
\end{center}

\paragraph{Phase~2 mixture.}
The $415$ composite demonstrations are combined with $560$ atomic replay demonstrations ($20$ for each task), totaling $975$ episodes. Replay contributes $127{,}150$ frames; the merged dataset contains $454{,}758$ frames. Training samples naturally by frame without class balancing, so composite and atomic replay account for $72.04\%$ and $27.96\%$ of the frame population.

\subsection{Real-Robot Setup}
\label{app:realrobot-setup}

\paragraph{Hardware.}
The platform is an AgileX PIPER $6$-DoF arm with a parallel-jaw gripper. Observations are RGB images from a single wrist-mounted fisheye camera, together with the arm's joint state and gripper aperture. The action expert produces chunks of horizon $H = 10$ at $30\,$Hz, giving each chunk a lookahead of $H/f \approx 0.33\,$s.

\paragraph{Tasks.}
We define two atomic skill families and one composite task.
\begin{itemize}
  \setlength{\itemsep}{0pt}
  \setlength{\parsep}{0pt}
  \item \textbf{Cup\,$\rightarrow$\,Table.} A cup is initially placed on top of a doll; the robot lifts the cup off the doll and places it on the table surface.
  \item \textbf{Doll\,$\rightarrow$\,Plate.} A doll is initially on the table; the robot picks up the doll and places it on a target plate.
  \item \textbf{Composite (Cup\,$\rightarrow$\,Table $\rightarrow$ Doll\,$\rightarrow$\,Plate).} The composite chains the two atomic skills sequentially.
\end{itemize}
The composite task requires a primitive switch mid-trajectory once the cup has been placed and the robot turns to approach the doll.  Each pick event corresponds to an open-to-close gripper edge in the action stream, providing an event anchor for the action-derived labeling oracle introduced in Section~\ref{sec:method:classifier}.

\begin{figure}[t]
  \centering
  \includegraphics[width=\linewidth]{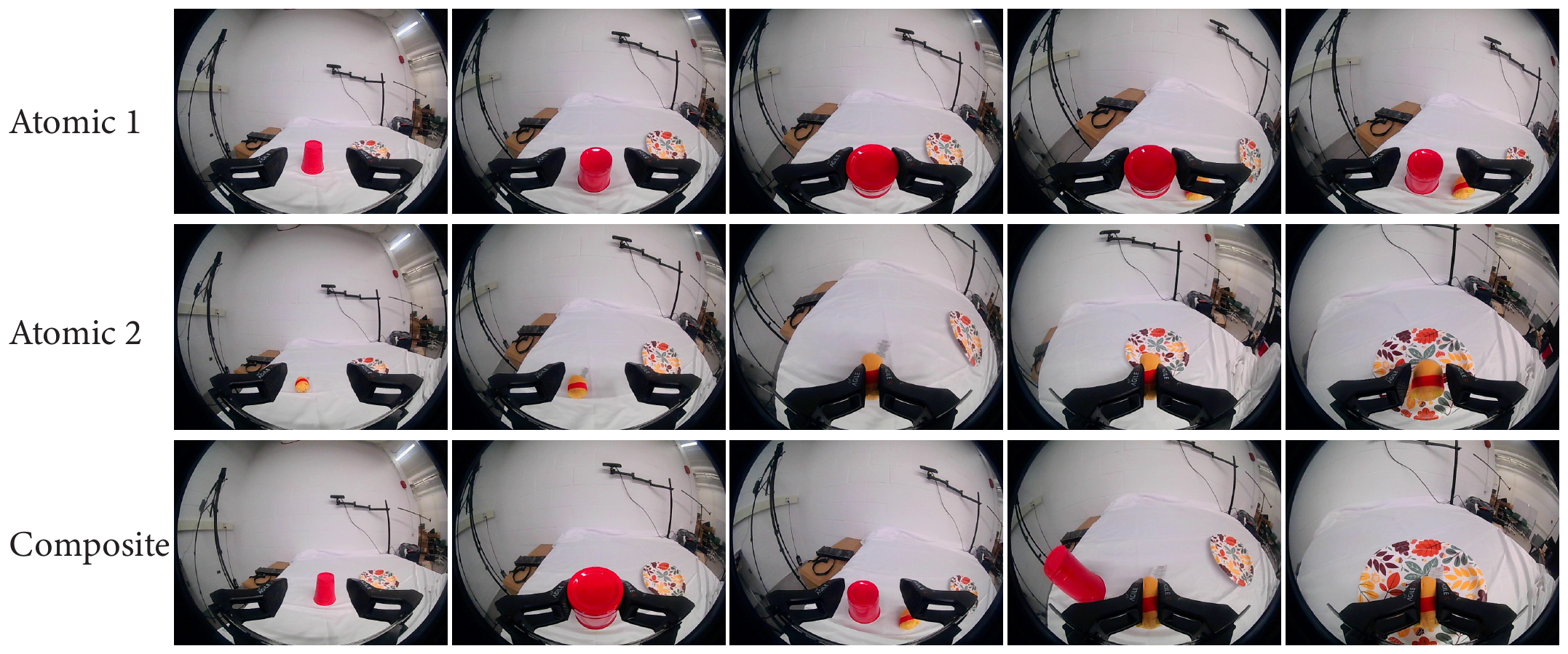}
  \caption{\textbf{Real-robot training data: fisheye observations.} Representative frames from the wrist-mounted fisheye RGB stream used for teleoperated training, sampled left-to-right along each trajectory. \textbf{Top:} \textit{Cup\,$\rightarrow$\,Table} (Atomic 1) -- the red cup is lifted off the doll and placed on the table. \textbf{Middle:} \textit{Doll\,$\rightarrow$\,Plate} (Atomic 2) -- the doll is picked up and placed on the patterned plate. \textbf{Bottom:} the \textit{Composite} chains the two atomic skills in sequence.}
  \label{fig:real-rob-data}
\end{figure}

\paragraph{Demonstrations and evaluation protocol.}
We collect $20$ teleoperated demonstrations per task ($60$ episodes total: $40$ atomic and $20$ composite). Each method is evaluated for $20$ trials per task using closely matched initial configurations. We report mean success rate over the $20$ trials; success is judged by a human evaluator under the criteria stated above (Cup\,$\rightarrow$\,Table / Doll\,$\rightarrow$\,Plate / Composite). We compare PRISM against the same four $\pi_{0.5}$ baselines as in simulation (zero-shot; LoRA atomic-only; LoRA two-stage atomic + composite; LoRA + replay). PRISM uses $N = 2$ primitive adapters and the kinematic labeling oracle, with atomic training followed by composite training.

\section{Kinematic Labeling Oracle: Full Rule}
\label{app:kinematic-oracle}

This appendix specifies the action-derived rule used to instantiate the labeling oracle $\mathcal{A}$ in Section~\ref{sec:method:classifier}.

\paragraph{Composite skill labeling rules.}
\label{app:a2-labels}
The labeler combines task identity with the gripper-command channel $g_t=a_t[11]$, treating $g_t>0.5$ as closed. Each composite demonstration is represented by two ordered skill phases separated by one transition. Multiple grasp-and-release interactions within a phase share the same skill label. Atomic demonstrations receive their task's skill label throughout.

\paragraph{Post-release transitions.}
LoadDishwasher and SnackSorting follow Pick\,\&\,Place $\to$ Close, while ThawInSink follows Pick\,\&\,Place $\to$ Turn\,Lever. Let $R$ be the last closed-to-open gripper event. The transition is placed at $T=R+30$, adding $1.5$ seconds at the $20$\,Hz demonstration rate before switching to the second skill.

\paragraph{Pre-grasp transitions.}
WashLettuce follows Turn\,Lever $\to$ Pick\,\&\,Place. Let $G$ be the last open-to-closed gripper event. The transition is shifted earlier according to $T=G-\min(210,0.35G)$, including the approach to the grasp in the Pick\,\&\,Place phase. The maximum backshift is $210$ frames ($10.5$ seconds), capped at $35\%$ of the elapsed frames preceding the grasp event.

\paragraph{Label assignment and supervision.}
Frames before the transition receive the first skill label, and the subsequent phase receives the second. If the required gripper event is absent, the entire demonstration receives the first skill label. Routing cross-entropy excludes frames within $100$ frames ($5$ seconds) of the transition, while all sampled frames contribute to flow matching. The labels supervise the router during training; composite-task inference uses its predicted soft weights.

\section{Vision-Language-Action Models}
\label{app:related-vla}

This appendix collects the broader VLA backbone literature. Vision-Language-Action (VLA) models combine pretrained vision-language representations with robot action prediction, enabling language-conditioned manipulation across diverse tasks~\citep{brohan2023rt1roboticstransformerrealworld, brohan2023rt2visionlanguageactionmodelstransfer, driess2023palmeembodiedmultimodallanguage, kim2024openvla, octomodelteam2024octoopensourcegeneralistrobot, black2024pi0visionlanguageactionflowmodel, intelligence2025pi05, nvidia2025gr00tn1openfoundation}. Recent systems improve this substrate through larger and more diverse robot data~\citep{embodimentcollaboration2025openxembodimentroboticlearning}; alternative action representations including autoregressive tokens, diffusion policies, flow matching, and chunked action prediction~\citep{pertsch2025fastefficientactiontokenization, chi2024diffusionpolicyvisuomotorpolicy, zhao2023learningfinegrainedbimanualmanipulation, zhao2024alohaunleashedsimplerecipe, liu2025rdt1bdiffusionfoundationmodel, liu2025hybridvlacollaborativediffusionautoregression}; and stronger spatial, hierarchical, or reasoning-augmented control modules~\citep{shi2025hirobotopenendedinstruction, wen2025dexvlavisionlanguagemodelplugin, lin2026onetwovlaunifiedvisionlanguageactionmodel, li2024cogactfoundationalvisionlanguageactionmodel, wen2025diffusionvlageneralizableinterpretablerobot, qu2025spatialvlaexploringspatialrepresentations, guo2025gladgeometriclatentdistillation, zhen20243dvla3dvisionlanguageactiongenerative, wen2025tinyvlafastdataefficientvisionlanguageaction}. These works primarily improve the VLA backbone, action decoder, data scale, or reasoning interface. In contrast, PRISM studies how a pretrained VLA should be \emph{adapted}: rather than absorbing each new behavior into one shared update, PRISM gives reusable skills parameter-level identity and learns to route among them during long-horizon execution.

\end{document}